\RequirePackage{fix-cm}
\documentclass[twocolumn]{svjour3}          
\smartqed  
\usepackage{graphicx}
\usepackage{times}
\usepackage{epsfig}
\usepackage{graphicx}
\usepackage{amsmath}
\usepackage{amssymb}
\usepackage{booktabs}
\usepackage{multirow}
\usepackage{subcaption}
\usepackage{mmstyle}
\usepackage[pagebackref=true,breaklinks=true,bookmarks=false,colorlinks]{hyperref}

\newcommand{\jm}[1]{\textcolor{black}{#1}}
\makeatletter
\DeclareRobustCommand\onedot{\futurelet\@let@token\@onedot}
\def\@onedot{\ifx\@let@token.\else.\null\fi\xspace}

\def\ie{\emph{i.e}\onedot} 
 
 \def\vs{\emph{vs}\onedot}
 
\def\etal{\emph{et al}\onedot}
\makeatother
\journalname{IJCV}
\begin{document}

\title{Towards Balanced Learning for Instance Recognition
\thanks{Code is available at \url{https://github.com/open-mmlab/mmdetection}.}
}


\author{Jiangmiao Pang$^{1}$ \and
        Kai Chen$^{2}$ \and
        Qi Li$^{1}$ \and
        Zhihai Xu$^{1}$ \and
        Huajun Feng$^{1}$ \and
        Jianping Shi$^{3}$ \and
        Wanli Ouyang$^{4}$ \and
        Dahua Lin$^{2}$
}


\institute{Jiangmiao Pang (pjm@zju.edu.cn) \\
           Kai Chen (ck015@ie.cuhk.edu.hk) \\
           Qi Li (liqi@zju.edu.cn) \\
           Zhihai Xu (xuzh@zju.edu.cn) \\
           Huajun Feng (fenghj@zju.edu.cn) \\
           Jianping Shi (shijianping@sensetime.com) \\
           Wanli Ouyang (wanli.ouyang@sydney.edu.au) \\
           Dahua Lin (dhlin@ie.cuhk.edu.hk) \\
           1 Zhejiang University \\
           2 The Chinese University of Hong Kong \\
           3 SenseTime Research \\
           4 The University of Sydney
}

\date{Received: date / Accepted: date}

\maketitle

\begin{abstract}
Instance recognition is rapidly advanced along with the developments of various deep convolutional neural networks.
Compared to the architectures of networks, the training process, which is also crucial to the success of detectors, has received relatively less attention.
In this work, we carefully revisit the standard training practice of detectors, and find that the detection performance is often limited by the imbalance during the training process, which generally consists in three levels -- sample level, feature level, and objective level.
To mitigate the adverse effects caused thereby, we propose Libra R-CNN, a simple yet effective framework towards balanced learning for instance recognition.
It integrates IoU-balanced sampling, balanced feature pyramid, 
and objective re-weighting, respectively for reducing the imbalance at sample, feature, and objective level.
Extensive experiments conducted on MS COCO, LVIS and Pascal VOC datasets prove the effectiveness of the overall balanced design.
\end{abstract}

\section{Introduction}
Instance recognition~\cite{det_review_ijcv,zou2019object}, a fundamental problem in computer vision, aims to automatically recognize what and where are the objects in the scenarios. 
Accurately and efficiently detecting objects is of vital importance for numerous downstream applications such as video analysis~\cite{otb,motreview,kang2016object,zhu2017flow} and autonomous driving~\cite{kitti,chen2017multi,chen2016monocular}.
Recently, various detection frameworks are proposed and substantially push forward the states of the art.
Those detectors can be mainly categorized into single-stage detectors such as RetinaNet~\cite{focalloss}, and two-stage detectors such as Faster R-CNN~\cite{frcnn}.

Despite the apparent differences in these architectures, modern detection frameworks mostly follow a common training paradigm, namely, sampling regions, extracting features therefrom,
and then jointly recognizing the categories and refining the locations under the guidance of a standard multi-task objective function.
\begin{figure}
	\setlength{\belowcaptionskip}{-15pt}
	\centering
	\includegraphics[width=\linewidth]{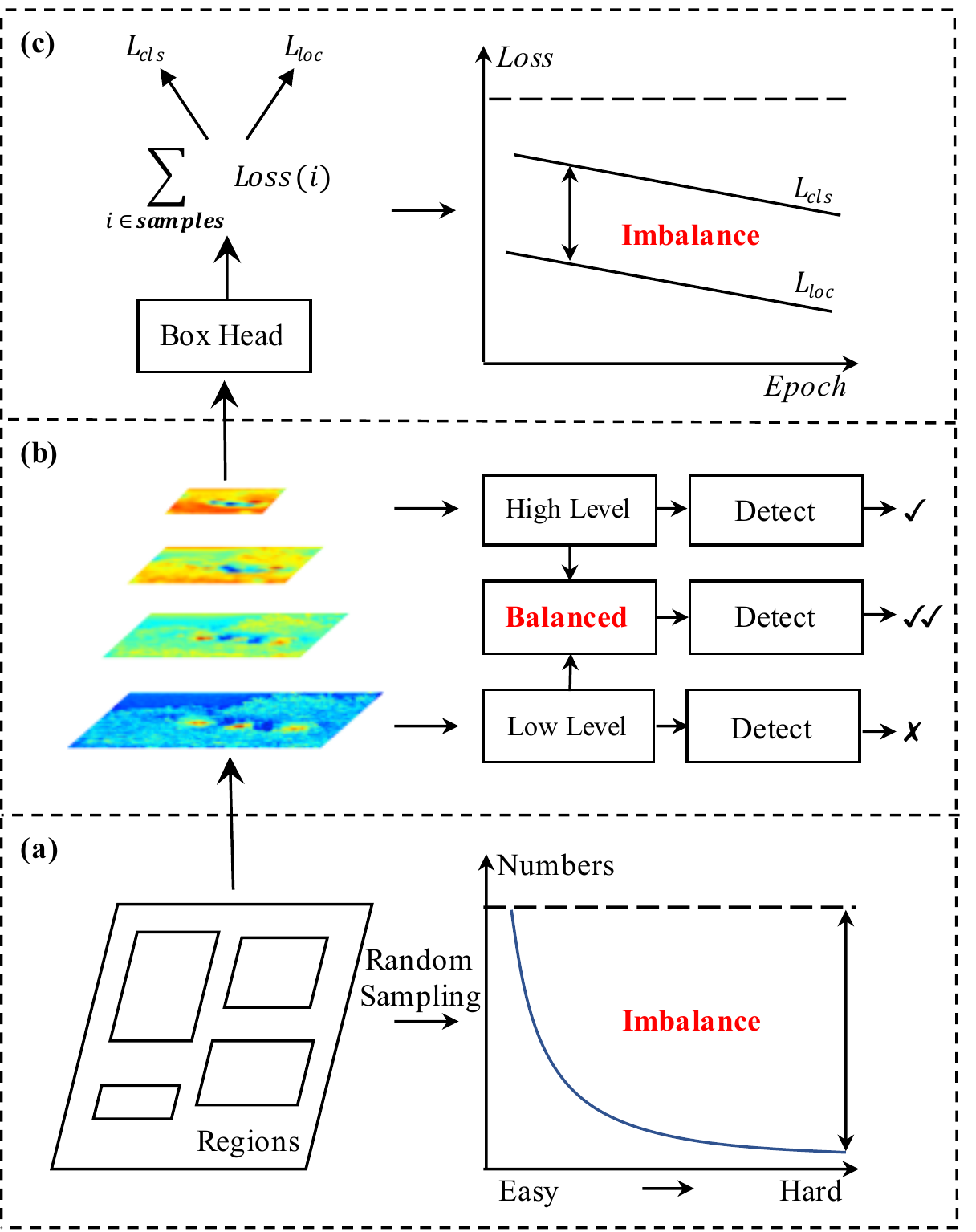}
	\caption{Imbalance consists in (a) sample level (b) feature level and (c) objective level,
	which prevents the well-designed model architectures from being fully exploited.}
	\label{fig:imbalance}
\end{figure}
%
Based on this common paradigm, obtaining high performance detectors not only relies on powerful model architectures but also on better training process.
However, we find that recent literature~\cite{cascadercnn,garpn,li2019zoom,cornernet_ijcv,zhang2019single} focus more on proposing better model architectures.
The training process, which is also crucial to the success of the detectors, has received relatively less attention in recent years. 

In this paper, we carefully revisit the standard training process of detectors, and argue that the key to successfully train the detectors mainly lies in three aspects:
(1) whether the selected region samples are representative,
(2) whether the extracted visual features are fully utilized, and
(3) whether the designed objective function is optimal.
Our study further reveals that the typical training process is significantly imbalanced in all these aspects, which prevents the power of the model architectures from being fully exploited so that limit the overall performance. 
Below, we describe these issues in turn and show them in Figure~\ref{fig:imbalance}:

1) Sample level imbalance:
Recent works often select pre-defined local regions in images, such as anchors~\cite{frcnn,focalloss,ssd,yolo} and proposals~\cite{rcnn,frcnn,fpn,maskrcnn}, as samples to train the detectors.
More representative the samples are, faster and better the detectors converge.
Obviously, hard samples, which the detectors cannot successfully recognize, could provide more cues to the training optimizer compared to easy samples.
However, the random sampling scheme usually results in
the selected samples dominated by easy ones.
Meanwhile, the random sampling may also bury the samples of rare classes under numerous frequent objects if there is an obvious class imbalance issue.

2) Feature level imbalance:
Deep high-level features in backbones are with more semantic meanings while the shallow low-level features are more content descriptive~\cite{zeiler2014visualizing}.
Recent feature integration through lateral connections like FPN~\cite{fpn} and PANet~\cite{panet} hava advanced the developments of instance recognition.
These methods inspire us that the low-level and high-level information are complementary.
The approaches how they are integrated to construct the feature pyramid matter to the following bounding box predictions.
As it is, what is the best approach to integrate them together?
Our study reveals that the integrated features should possess balanced information from different feature levels.
Especially when the features forwarded to the heads are extracted from different levels~\cite{fpn,focalloss,maskrcnn,cascadercnn}. 
In this case, the input region features may have different representations thus increasing the training difficulty. 
However, the sequential manner in aforementioned methods will make integrated features focus more on adjacent resolution but less on others.
The semantic information contained in non-adjacent levels would be diluted once per fusion during the sequential flow that makes the feature pyramids bias on different levels.

3) Objective level imbalance:
Objective functions measure gradients from the output features of training samples and optimize the model weights accordingly.
Since the training samples span from easy to hard and distribute across different classes, the measured gradients are also imbalanced in these aspects. 
If the gradients of easy and hard samples are not properly balanced, the small gradients produced by the easy samples may be drowned into the large gradients from the hard ones.
Similarly, the classifier in the model mostly receives gradients from samples of frequent classes. 
Due to the lack of enough positive samples~\cite{tan2020equalization}, the corresponding weights for classifying rare classes mostly receive discouraging gradients from the other classes.


To mitigate the adverse effects caused by these issues,
we propose Libra R-CNN, a simple yet effective framework for instance recognition that
explicitly enforces the balance at all three levels discussed above.
This framework integrates:
(1) \emph{IoU-balanced sampling}, which mines hard samples according to their IoU with assigned ground-truth.
We also uniformly select samples for different instances or classes as complementary.
(2) \emph{balanced feature pyramid},
which strengthens the multi-level features using the same deeply integrated balanced semantic features.
(3) \emph{objective re-weighting}, which use logarithmic gradient re-weighting to balance gradients from different samples, and bucketed class re-weighting to balance gradients for different classes. The conclusions also show that the re-weighting design in loss functions should joint consider the distribution of training samples.


We conducted extensive experiments on MS COCO~\cite{coco}, LVIS~\cite{lvis}, and Pascal VOC~\cite{voc} to prove the effectiveness of the proposed method.
With the overall balanced design, Libra R-CNN obtains 3.0 points and 2.0 points higher Average Precision (AP) than FPN Faster R-CNN and RetinaNet respectively on MS COCO~\cite{coco}.
Besides, Libra R-CNN achieves 43.7 and 47.2 points AP with ResNet-101 and ResNext-64x4d-101-DCN backbone on COCO \emph{test-dev} subset respectively, which are on par with recent states of the art.
When extending to LVIS dataset, our method also outperforms Mask R-CNN baseline by 3.0 points on overall AP and 6.1 points on rare classes.

Preliminary results of this work have been published in Pang \etal~\cite{librarcnn}.
The current work has been improved and extended to the journal version in several aspects.
(1) We analyze and propose an equivalent variant of balanced $L_1$ loss called Logarithmic Gradient Re-weighting(LGR). We adopt LGR to cross-entropy loss and observe consistent improvements. The results can prove that the re-weighting loss function design like Focal Loss are related to the distribution of training samples.
(2) We propose bucketed class re-weighting that tackles the class imbalance issue from the objective aspect. On LVIS dataset, the method outperforms Mask R-CNN baseline by 5.8 points and EQL~\cite{tan2020equalization} by 2.0 points on the AP of rare classes.
(3) More experiments are conducted on LVIS and Pascal VOC datasets, and extended from object detection to instance segmentation to prove the effectiveness of the proposed method. 
(4) We rewrite and polish the whole paper to make it more readable.

\section{Related Work}
In this section, we first talk about detectors with different model architectures, which dominate recent advancements of instance recognition. Then we talk about the related works that balance the detectors in different aspects.

\subsection{Model architectures for instance recognition.}
Model architectures in object detectors can be categorized into two-stage detectors and single-stage detectors.

Two-stage detectors are first introduced by R-CNN~\cite{rcnn} that classifies the proposals generated by Selective Search~\cite{selectivesearch} with the aid of CNNs~\cite{alexnet} and SVM~\cite{svm}.
SPP~\cite{spp} and Fast R-CNN~\cite{fastrcnn} are then proposed with spatial pyramid pooling and RoI Pooling to boost the performance.
Faster R-CNN~\cite{frcnn}, a meaningful milestone in instance recognition, introduces Region Proposal Network~(RPN) to generate proposals and allow the entire detectors to be end-to-end trainable.
After the milestone, numerous methods are proposed to boost Faster R-CNN from different aspects.
For example, R-FCN~\cite{rfcn} proposes position-sensitive RoI pooing and make the detector fully convolutional.
FPN~\cite{fpn} constructs a feature pyramid with top-down lateral connections and tackle the scale variance problem.
Cascade R-CNN~\cite{cascadercnn} extends the two-stage Faster R-CNN to a multi-stage manner with the classic yet powerful cascade architecture.
Mask R-CNN~\cite{maskrcnn} extends Faster R-CNN by adding a mask branch that refines the detection results under the help of multi-task learning.
\jm{Mask Scoring R-CNN~\cite{huang2019mask} and BMask R-CNN~\cite{cheng2020boundary} further improves the performance of Mask R-CNN by learning the quality of the predicted instance masks and leveraging object boundary information to improve mask localization accuracy respectively.}
HTC~\cite{htc} improves the mask information flow in Mask R-CNN and improves the overall performance by hybrid task joint training.
These two-stage detectors are known by their high accuracy but are not efficient enough, thus single-stage detectors are proposed to improve the efficiency. 

The high efficiency single-stage detectors are first popularized by YOLO~\cite{yolo,yolo9000} and SSD~\cite{ssd}.
They are simpler and faster than two-stage detectors but have trailed the accuracy until the introduction of RetinaNet~\cite{focalloss}.
CornerNet~\cite{cornernet_ijcv} introduces an insight that the bounding boxes can be predicted as a pair of key points.
This idea is followed up by recent literature such as ExtremeNet~\cite{extremenet}, CenterNet~\cite{centernet} and RepPoints~\cite{reppoints}.
The accuracy of single-stage detectors is gradually boosted by these powerful architectures, especially the recent FCOS~\cite{fcos} and ATSS~\cite{atss}, even on par with the two-stage ones now.

Other frameworks focus on
	cascade procedures~\cite{ouyang2017chained},
	duplicate removal~\cite{relationnetwork,learningnms},
	multi-scales~\cite{cai2016unified,sniper2018,snip},
	adversarial learning~\cite{adaptdet}
and more contextual~\cite{zeng2018crafting}.
All of them try to boost detection performance with specific architectures on specific stages of the detection pipelines.

\subsection{Balanced learning for instance recognition.}
Alleviating imbalance in the training process of instance recognition is crucial to achieve an optimal training and fully exploit the potential of model architectures.

\paragraph{Sample level imbalance:}
OHEM~\cite{ohem} and focal loss~\cite{focalloss} are primary existing solutions for sample level imbalance in instance recognition.
The commonly used OHEM automatically selects hard samples according to their confidences.
However, this procedure causes extra memory and speed costs, making the training process bloated.
Moreover, the OHEM also suffers from noise labels so that it cannot work well in all cases.
Focal loss solved the extra foreground-background class imbalance in single-stage detectors with an elegant loss formulation,
but it generally brings little or no gain to two-stage detectors because of the different imbalanced situation.
Similar to focal loss, Gradient Harmonizing Mechanism (GHM)~\cite{ghm} is recently proposed to balance the easy and hard samples with sample number counting, and show that focal loss is the special case of them.
Besides, Gupta~\etal~\cite{lvis} propose repeat factor sampling that gives larger sampling weights to images of rare categories.
Compared to these methods, our method is substantially lower cost, and tackles the problem elegantly.

\begin{figure*}
	\centering
	\includegraphics[width=\linewidth]{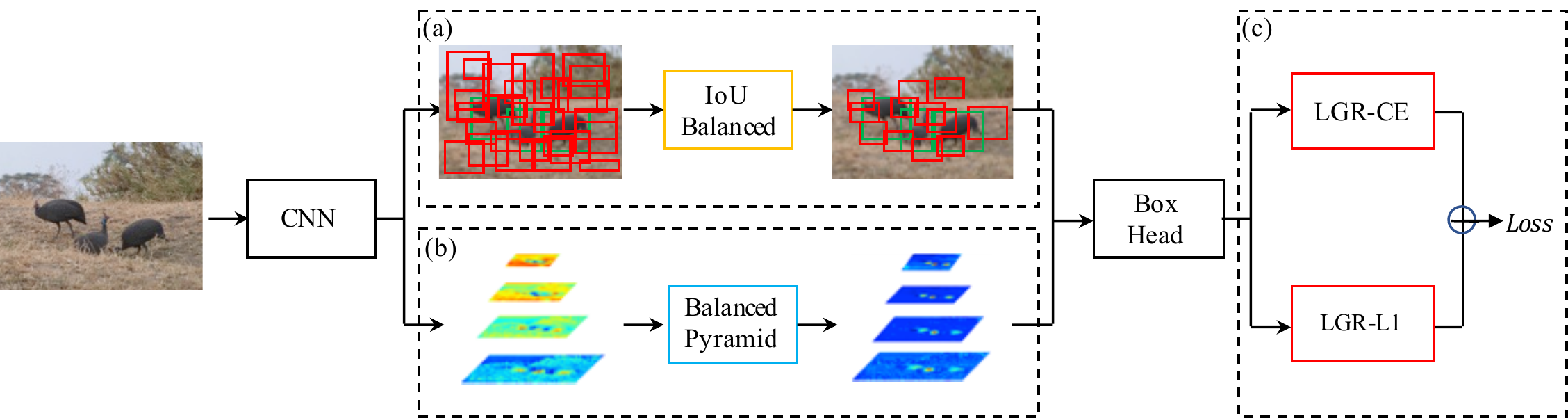}
	\caption{Overview of the proposed Libra R-CNN:
		an overall balanced design for instance recognition that integrates novel components (a) IoU-balanced sampling (b) balanced feature pyramid and (c) objective re-weighting respectively for reducing the imbalance at sample, feature, and objective level.}
	\label{fig:overall}
\end{figure*}

\paragraph{Feature level imbalance:}
More discriminative of the feature representations, better the recognition capability of the detector.
Therefore, how to integrate multi-level features in backbone to construct feature pyramids is a crucial problem for instance recognition.
Only using the features in shallow layers is incapable to do instance recognition until the introduction of FPN~\cite{fpn}. 
FPN~\cite{fpn} exploits lateral connections to enrich the semantic information of shallow layers from deep layers through a top-down pathway. 
With the feature pyramid, the bounding box head crops region features from different resolutions thus tackle the scale-invariance problem at the same time.
Inspired by the great work, PANet~\cite{panet} uses a bottom-up pathway to further increase the low-level information in deep layers as complementary.
Kong \etal~\cite{kong2018deep} proposes a novel efficient pyramid based on SSD that integrates the features in a highly-nonlinear yet efficient way.
Parallel FPN~\cite{kim2018parallel} employs the last layer of the backbone to generate feature pyramids with SPP~\cite{spp} to let different feature levels with similar semantic meanings. 
M2Det~\cite{m2det} continue extends the idea and build stronger
feature pyramid representations by employing multiple U-shape modules after a backbone model.
Recent, Google uses Neural Architecture Search (NAS) to search the best integration method to combine the multi-scale features and propose NAS-FPN~\cite{fpn}.
In contrast to these methods, our approach relies on integrated balanced semantic features to strengthen original features.
In this manner, each resolution in the pyramid obtains equal information from others, thus balancing the information flow and leading the features more discriminative.

\paragraph{Objective level imbalance:}
The aforementioned focal loss and GHM loss also can be treated as improving the objective level imbalance for different samples as the two problems are tightly related.
Besides, tackling class level imbalance from the objective perspective are mainly discussed on image classification~\cite{cao2019learning,cui2019class,khan2017cost,shen2016relay,wang2017learning,kang2019decoupling}.
After the release of LVIS dataset, 
To further tackle the class imbalance issue, Tan~\etal propose EQL~\cite{tan2020equalization} that ignores discouraging gradients for negative samples. 
Li~\etal propose balanced group softmax~\cite{li2020overcoming} that balances the classifiers through group-wise training.
In this paper, we propose bucketed class re-weighting that adaptively partitions the categories into different magnitudes, and use sigmoid function to adjust their weights.

\section{Methodology}

The overall pipeline of Libra R-CNN is shown in Figure~\ref{fig:overall}.
Our goal is to alleviate the imbalance exists in the training process of detectors using an overall balanced design,
thus exploiting the potential of model architectures as much as possible.
All components will be detailed in the following sections.

\begin{figure}
	\centering
	\includegraphics[width=\linewidth]{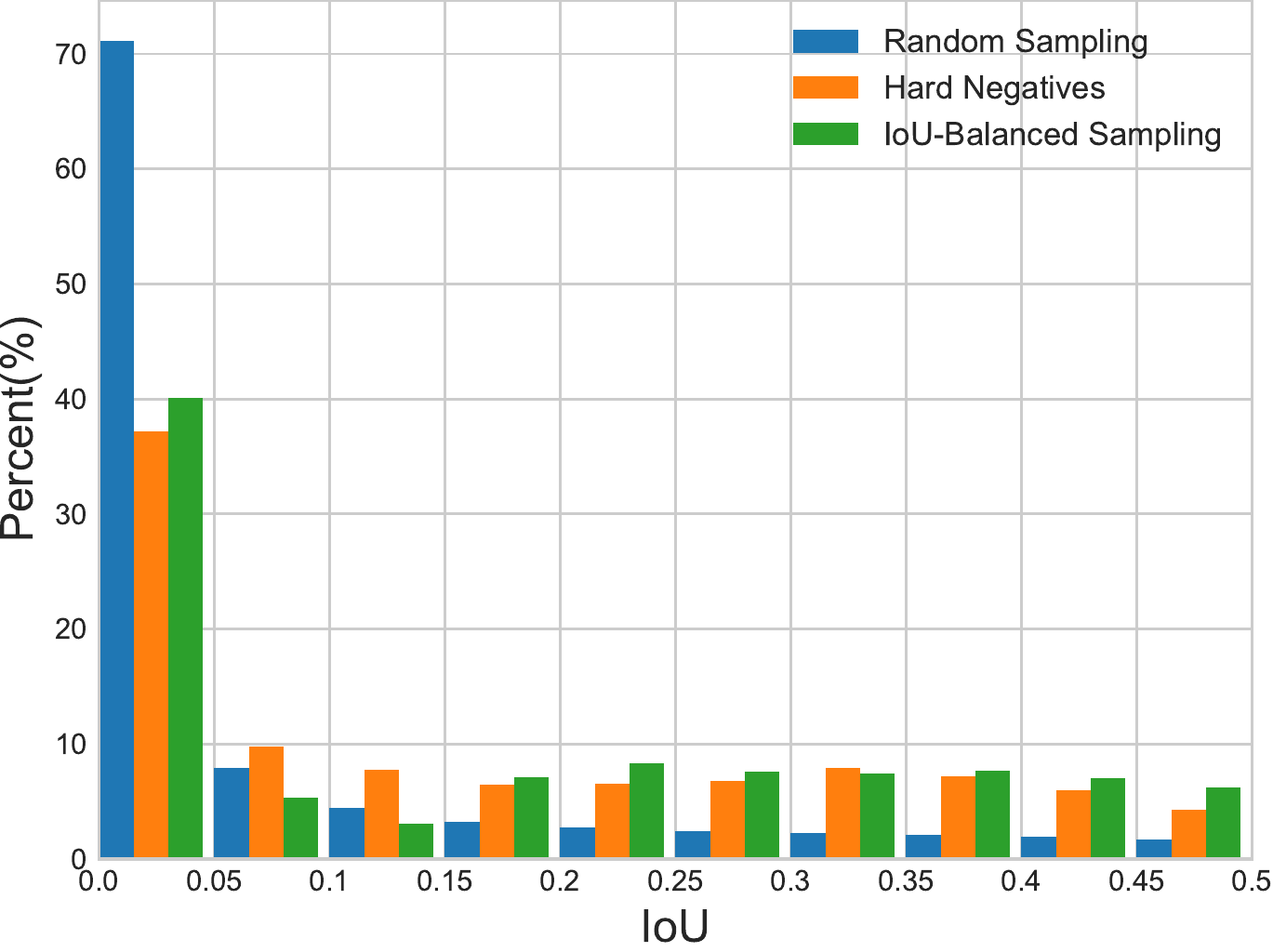}
	\caption{IoU distribution of random selected samples, IoU-balanced selected samples, and hard negatives.}
	\label{fig:sample_dist}
\end{figure}

\begin{figure*}
	\centering
	\includegraphics[width=\linewidth]{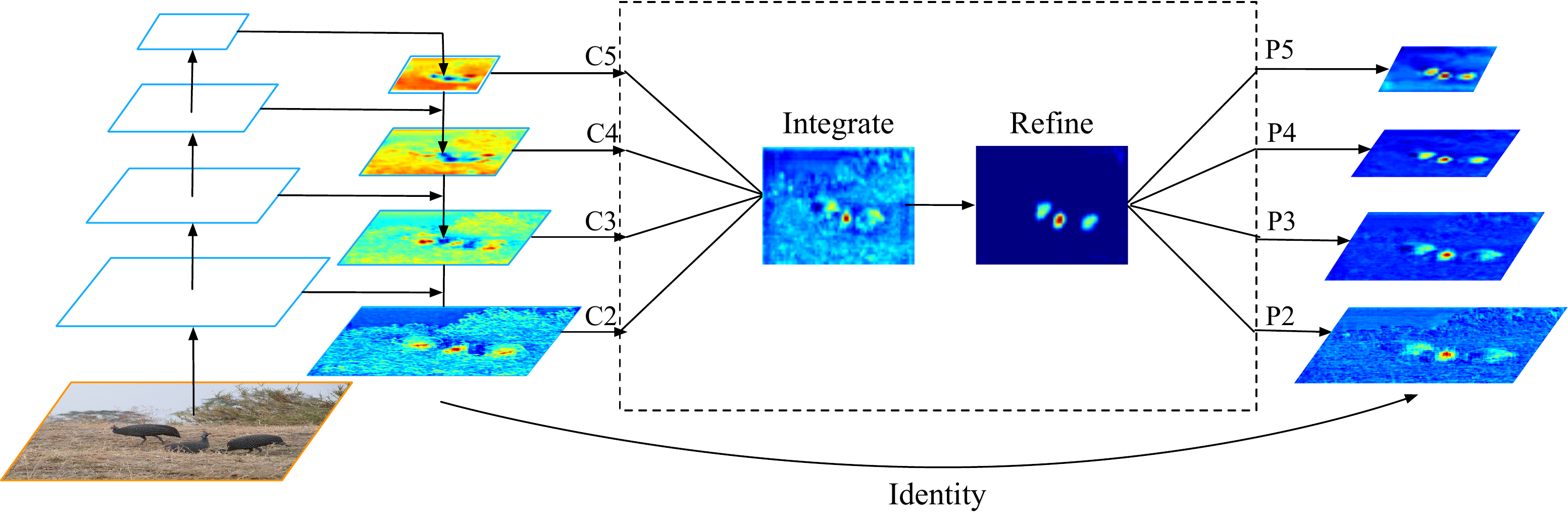}
	\caption{Pipeline and heatmap visualization of balanced feature pyramid.}
	\label{fig:fpn}
\end{figure*}

\subsection{IoU-balanced Sampling}
\label{subsec:sample}
Let us start with the basic question:
is the overlap between a training sample and its corresponding ground truth associated with its difficulty?
To answer this question, we conduct experiments to find the truth behind.
Results are shown in Figure~\ref{fig:sample_dist}.
We mainly consider hard negative samples, which are known to be the main problem.
We find that more than $60\%$ hard negatives have an overlap greater than $0.05$,
but random sampling only provides us $30\%$ training samples that are greater than the same threshold.
This extreme sample imbalance buries many hard samples into thousands of easy samples.

Motivated by this observation, we propose IoU-balanced sampling: a simple but effective hard mining method without extra cost.
Suppose we need to sample $N$ negative samples from $M$ corresponding candidates.
The selected probability for each sample under random sampling is
\begin{equation}
	\label{equ:random}
	p = \frac{N}{M}.
\end{equation}

To raise the selected probability of hard negatives,
we evenly split the sampling interval into $K$ bins according to IoU.
$N$ demanded negative samples are equally distributed to each bin.
Then we select samples from them uniformly.
Therefore, we get the selected probability under IoU-balanced sampling
\begin{equation}
	\label{equ:sample}
	p_{k} = \frac{N}{K} *\frac{1}{M_{k}}, ~~ k \in [0, K),
\end{equation}
where $M_{k}$ is the number of sampling candidates in the corresponding interval denoted by k.
K is set to 3 by default in our experiments.

The sampled histogram with IoU-balanced sampling is shown by green color in Figure~\ref{fig:sample_dist}.
It can be seen that our IoU-balanced sampling can guide the distribution of training samples close to the one of hard negatives.
Experiments also show that the performance is not sensitive to K,
as long as the samples with higher IoU are more likely selected.

Besides, it is also worth noting that the method is also suitable for hard positive samples.
However, in most cases, there are not enough sampling candidates to extend this procedure into positive samples.
To make the balanced sampling procedure more comprehensive,
we sample an equal number of positive samples for each ground truth as an alternative method.
When there is a severe class-level imbalance in the training dataset,
we first sample an equal number of positive samples for each class, then adopt the instance balanced sampling to samples of each class respectively.
This method can excavate enough training samples for rare classes nor they will be buried under numerous samples of frequent classes.

\subsection{Balanced Feature Pyramid}
In contrast to former approaches~\cite{fpn,panet} that integrate multi-level features using lateral connections,
our key idea is to \emph{strengthen} the multi-level features using the \emph{same} deeply integrated balanced semantic features.
The pipeline is shown in Figure~\ref{fig:fpn}.
It consists of four steps, rescaling, integrating, refining and strengthening.

\begin{figure*}
	\centering
	\includegraphics[width=\linewidth]{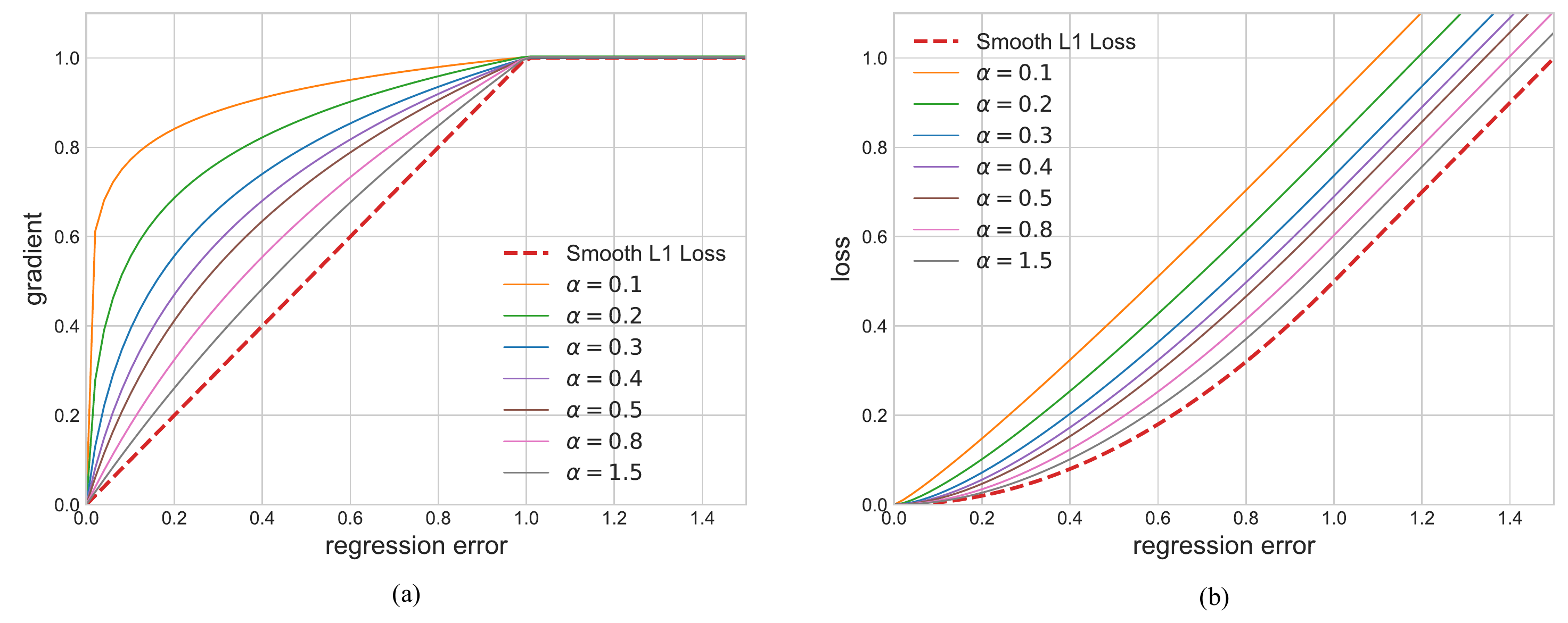}
	\caption{We show curves for (a) gradient and (b) loss of our balanced $L_1$ loss here. Smooth $L_1$ loss is also shown in dashed lines. $\gamma$ is set default as 1.0.}
	\label{fig:loss}
\end{figure*}

\paragraph{Obtaining balanced semantic features.}
Features at resolution level $l$ are denoted as $C_l$.
The number of multi-level features is denoted as L.
The indexes of involved lowest and highest levels are denoted as $l_{min}$ and $l_{max}$.
In Figure~\ref{fig:fpn}, $C_2$ has the highest resolution.
To integrate multi-level features and preserve their semantic hierarchy at the same time, we first resize the multi-level features $\{C_{2}, C_{3}, C_{4}, C_{5}\}$ to an intermediate size, \ie, the same size as $C_{4}$, with interpolation and max-pooling respectively.
Once the features are rescaled, the \emph{balanced semantic features} \textbf{f} are obtained by simple averaging as
\begin{equation}
	\label{equ:IDFeat}
	\textbf{f} = \frac{1}{L}\sum_{l=l_{min}}^{l_{max}}{C_{l}}.
\end{equation}
The obtained features are then rescaled using the same but reverse procedure to strengthen the original features.
Each resolution obtains equal information from others in this procedure.
Note that this procedure does not contain any parameter.
We observe improvement with this nonparametric method, proving the effectiveness of the information flow.

\paragraph{Refining balanced semantic features.}
The balanced semantic features can be further refined to be more discriminative.
We found both the refinements with convolutions directly and the non-local module~\cite{nonlocal} work well.
But the non-local module works more stable.
Therefore, we use the embedded Gaussian non-local attention as default in this paper.
That is, the features at each position in the integrated features $\textbf{f}_i$ are refined by
\begin{equation}
	\label{equ:nonlocal}
	\textbf{f}^{'}_i = \sum_{\forall j} \frac{ e^{\theta(\textbf{f}_i)^T \phi(\textbf{f}_j)}}{\sum_{\forall j} e^{\theta(\textbf{f}_i)^T \phi(\textbf{f}_j)}} g(\textbf{f}_j) + \textbf{f}_i,
\end{equation}
where j denotes all other positions in the feature maps and $\theta(\cdot)$, $\phi(\cdot)$, $\text{g}(\cdot)$ denote normal convolutions with a kernel size of $1 \times 1$.
The refining step helps us enhance the integrated features and further improve the results.

With this method, features from low-level to high-level are aggregated at the same time.
The outputs $\{P_{2}, P_{3}, P_{4}, P_{5}\}$ are used for instance recognition following the same pipeline in FPN.
It is also worth mentioning that our balanced feature pyramid can work as complementary with recent solutions such as FPN and PAFPN without any conflict.

\subsection{Objective Re-weighting}
Classification and localization problems are solved simultaneously under the guidance of a multi-task loss since Fast R-CNN~\cite{fastrcnn}, which is defined as
\begin{equation}
	\label{equ:multi-task}
	L_{p, u, t^u, v} = L_{cls}(p, u) + \lambda[u \ge 1]L_{loc}(t^u, v).
\end{equation}
$L_{cls}$ and $L_{loc}$ are objective functions corresponding to recognition and localization respectively.
Predictions and targets in $L_{cls}$ are denoted as $p$ and $u$.
$t^u$ is the corresponding regression results with class $u$.
$v$ is the regression target.
$\lambda$ is used for tuning the loss weight under multi-task learning.

\subsubsection{Balanced $L_1$ Loss}

For the regression objective function, we define samples with a loss greater than or equal to 1.0 outliers.
The other samples are called inliers.
The outliers, which can be regarded as hard samples, produce relatively larger gradients to inliers.
In this case, the inliers, which can be regarded as the easy samples, will contribute little gradients to the overall gradients.
To be more specific, inliers only contribute 30\% gradients average per sample compared with outliers.
Considering these issues, we propose balanced $L_1$ loss, which is denoted as $L_{b}$.

Balanced $L_1$ loss is derived from the conventional smooth $L_1$ loss,
in which an inflection point is set to separate inliers from outliners, and clip the large gradients produced by outliers with a maximum value of 1.0, as shown by the dashed lines in Figure~\ref{fig:loss}-(a).
The key idea of balanced $L_1$ loss is promoting the crucial regression gradients,
\ie gradients from inliers (accurate samples), to rebalance the involved samples, thus achieving a more balanced training.
Localization loss $L_{loc}$ uses balanced $L_1$ loss is defined as
\begin{equation}
	\label{equ:loc}
	L_{loc} = \sum_{i\in\{x,y,w,h\}}{L_{b}(t_{i}^{u} - v_i)},
\end{equation}
and its corresponding formulation of gradients follows
\begin{equation}
	\label{equ:partial}
	\frac{\partial{L_{loc}}}{\partial{w}} \varpropto
	\frac{\partial{L_{b}}}{\partial{t_i^u}} \varpropto
	\frac{\partial{L_{b}}}{\partial{x}},
\end{equation}
Based on the formulation above, we design a promoted gradient formulation as
\begin{equation}
	\label{equ:gradient}
	\frac{\partial{L_{b}}}{\partial{x}} =
	\begin{cases}
		\alpha ln(b|x| + 1) & \text{if $|x| < 1$} \\
		\gamma  & \text{otherwise},
	\end{cases}
\end{equation}

Figure~\ref{fig:loss}-(a) shows that our balanced $L_1$ loss increases the gradients of inliers under the control of a factor denoted as $\alpha$.
A small $\alpha$ increases more gradient for inliers, but the gradients of outliers are not influenced.
Besides, an overall promotion magnification controlled by $\gamma$ is also brought in for tuning the upper bound of regression errors, which can help the objective function better balancing involved tasks.
The two factors that control different aspects are mutually enhanced to reach a more balanced training.
$b$ is used to ensure $L_{b}(x=1)$ has the same value for both formulations in Eq. (\ref{equ:loss}).

By integrating the gradient formulation above, we can get the balanced $L_1$ loss
\begin{equation}
	\label{equ:loss}
	L_{b}(x) =
	\begin{cases}
		\frac{\alpha}{b}(b|x| + 1)ln(b|x| + 1) - \alpha |x|  & \text{if $|x| < 1$}\\
		\gamma|x| + C  & \text{otherwise},
	\end{cases}
\end{equation}
in which the parameters $\gamma$, $\alpha$, and $b$ are constrained by
\begin{equation}
	\alpha ln(b+1) = \gamma.
\end{equation}
The default parameters are set as $\alpha = 0.5$ and $\gamma = 1.5$ in our experiments.

\subsubsection{Logarithmic Gradient Re-weighting}

\paragraph{Variant of balanced $L_1$ loss:}
The key idea of balanced $L_1$ loss is using Eq.~\ref{equ:gradient} to increase the gradients of inliers.
In this manner, the optimization procedure of the network could have larger momentum when the model is good enough but not great enough, \ie, capable to optimize the small errors.
Therefore, the loss function can especially enhance the accurate localization capability when the proposals are close to the ground truth, which is shown in the experimental part.

In this section, we introduce another simple equivalent variant of balanced $L_1$ loss, as a complementary of the original Libra R-CNN design~\cite{librarcnn}.

We denote the loss function and the gradient formulation of smooth $L_1$ loss as $L_s$ and $G_s$ respectively:
\begin{equation}
	\label{equ:smoothl1}
	L_s =
	\begin{cases}
		0.5x^2 & \text{if $|x| < 1$}, \\
		|x|-0.5  & \text{otherwise},
	\end{cases}
\end{equation}

\begin{equation}
	\label{equ:smoothl1grad}
	G_s = \frac{\partial{L_s}}{\partial{x}} =
	\begin{cases}
		|x| & \text{if $|x| < 1$}, \\
		1  & \text{otherwise}.
	\end{cases}
\end{equation}

We also denote the value of Eq.~\ref{equ:gradient} as $G_b$.
The promoted ratio $\rho$ between the gradients of two loss function is:
\begin{equation}
	\label{equ:ratio}
	\rho = \frac{G_b}{G_s} =
	\begin{cases}
		\frac{\alpha ln(b|x| + 1)}{|x|} & \text{if $|x| < 1$}, \\
		\gamma.
	\end{cases}
\end{equation}

During model optimization, we \emph{detach} the value $\rho$ from gradient calculation thus it could be treated as a constant value.
Then we can obtain the variant of $L_b$ as
\begin{equation}
	\label{equ:smoothl1v}
	L_b = \rho L_s =
	\begin{cases}
		0.5 \rho x^2 & \text{if $|x| < 1$}, \\
		\rho(|x|-0.5)  & \text{otherwise}.
	\end{cases}
\end{equation}

We call this adaptive gradient-level re-weighting method Logarithmic Gradient Re-weighting(LGR).

\paragraph{LGR for Cross-entropy Loss:}
Cross-entropy loss is one of the most popular classification loss and well used in image classification, object detection and instance segmentation.
Here we directly talk about classification with multiple classes, that is, the cross-entropy loss with softmax.
Assume there are C classes in the dataset and we generally predict a vector \textbf{z} whose dimension is $C+1$ to obtain the probability with softmax:
\begin{equation}
    \label{equ:softmax}
    p_i = \frac{e^{z_i}}{\sum_{c=0}^C e^{z_c}}, ~~~~~ \text{for i = 0 ... C}.
\end{equation}
The cross-entropy loss function follows
\begin{equation}
    \label{equ:ce}
    L(p, t) = -\sum_{c=0}^C t_c * log(p_c),
\end{equation}
where $t$ is 1 when c equals with the ground truth label else 0.

Then we can get the gradient function of cross-entropy loss with softmax:
\begin{equation}
    \label{equ:ce_g}
    \frac{\partial{L(p, t)}}{\partial{z_c}} = p_c - t_c.
\end{equation}
We use the norm of the gradients of corresponding samples to calculate the weights after LGR.
Hence,
\begin{equation}
	\label{equ:ratio2}
	\rho_c =
		\frac{\alpha ln(b * \lvert p_c - t_c \rvert + 1)}{\lvert p_c - t_c \rvert}
\end{equation}
The cross-entropy loss becomes
\begin{equation}
    \label{equ:ce2}
    L(p, t) = -\sum_{c=0}^C \rho_c t_c * log(p_c),
\end{equation}

We observe consistent improvements when exploiting this loss function.
It is interesting that the formulation has an opposite conclusion \vs Focal Loss~\cite{focalloss}, which down-weights the losses of easy samples.
We conclude this phenomenon related to the distribution of training samples.

In single-stage detectors like SSD / RetinaNet, there are $\sim$100k locations that densely over spatial positions.
All locations contribute their gradients to the training process without sampling procedure.
Hence, even the small gradients the easy samples generated, the large number still make the overall gradients of easy samples dominating.
The focal loss is thus proposed to down-weight the loss assigned to well-classified examples.

In contrast, for two-stage detectors like Faster R-CNN, the common training settings usually use 512 samples for training.
The number of easy samples has a comparable number with hard samples.
Hence, the easy number may contribute little to the overall gradients along with the optimization procedure.
This conclusion also explains that why Focal Loss cannot work well on two-stage detectors.
The conclusion here proves that different sample distributions need different re-weighting methods.
As the training samples need to be representative, the designed loss function should well-maintain different sample types.
The key point is to obtain an overall gradient balancing between different sample types.

\subsubsection{\jm{Bucketed} Class Re-weighting}
If there is an obvious class imbalance in the training dataset,
obtaining robust classifiers for rare classes will be much more challenging.
The final performance of different classes is highly relevant to the number of training samples~\cite{lvis}.
For the classifier, the weights for classifying frequent classes are adequately trained with enough positive samples.
However, due to the limited positive samples of rare classes, the weights for classifying rare classes receive too many discouraging gradients from samples of the other classes~\cite{tan2020equalization}.
Therefore, for a training sample, we can decrease its gradients propagated to non-related negative rare classes except the background one to alleviate the imbalance issue.

Assume we are using cross entropy loss with sigmoid to classify multiple classes for simplicity and there are $\mathcal{C}$ classes in the dataset.
For a training sample whose class is $t$, the classification loss is
\begin{equation}
    L_c = - \sum^{\mathcal{C}}_{i=0} w_i log(p_i),
\end{equation}
where $p_i$ denotes the classification probability of corresponding class, and $i = 0$ denotes the background. 
We set $w_i$ as 1.0 when i equals to 0 or $t$ by default.
However, the frequencies of different classes in the training dataset do not have obvious rules and most have large variances.
As it is, how to set the weights for the non-related negative classes?

In this paper, we use bucketed class re-weighting to tackle the issue.
We first uniformly bucketing the frequencies of the classes into a normalized vector, and then use sigmoid function to obtain the adjusting weights.

Specifically, considering each class has a frequency of $f_i$ in the training dataset, 
As the frequencies of different classes may have large variance, we use 
\begin{equation}
    f^{'}_i = ln(f_i)
\end{equation}
to decrease the variance first. 
Assume we hope to split the classes into $S$ bins, we evenly partition $f^{'}$ into $S$ ranges according to its maximum and minimum values.
The bin index $s_i$ for each class are obtained if $f^{'}_i$ falls into corresponding range, where $s_i$ = 0, ..., $S - 1$.
Then we normalize the bin indexes follows
\begin{equation}
    \textbf{x}_i = \frac{s_i}{S - 1}, ~~~~~ i~\in~[0, S).
\end{equation}
With the bucketing operation, the classes are uniformly distributed and normalized to a vector $\textbf{x}$.

Then we use sigmoid function to obtain the weight for each class following
\begin{equation}
    \label{equ:sigmoid}
    w_i = \frac{1}{1 + e^{-(\textbf{x}_i-a)/t}},
\end{equation}
where $t$ is the temperature parameter and $a$ is use to set the turning point where mostly turn at common classes.
Obviously, sigmoid function is naturally to adjust the weights for vector $\textbf{x}$.
As $\textbf{x}$ is already symmetrical for the class frequencies, it can accordingly give symmetrical large weights to frequent classes and small weights to rare classes, and turning at common classes.
We set $S$ as 10, $a$ as 0.5 and $t$ as 0.025 in this paper by default.

\section{Experiments}

We conduct experiments on MS COCO~\cite{coco} and Pascal VOC~\cite{voc} to verify the effectiveness of the proposed method.
We also conduct experiments on Large Vocabulary Instance Segmentation (LVIS)~\cite{lvis} to verify the effectiveness of our efforts on class-level balancing.

\begin{table*}[htb]
	\centering
	\addtolength{\tabcolsep}{1pt}
	\begin{tabular}{*{12}{c}}
		\toprule
		Method                       & Backbone  & AP & $\text{AP}_{50}$ & $\text{AP}_{75}$ & $\text{AP}_{S}$ & $\text{AP}_{M}$ & $\text{AP}_{L}$  \\
		\midrule
        YOLOv2~\cite{yolo9000} & DarkNet-19 &21.6 &44.0 &19.2 &5.0 &22.4 &35.5\\
        SSD512~\cite{ssd} & VGG-16 &28.8 &48.5 &30.3 &10.9 &31.8 &43.5 \\
        RetinaNet~\cite{focalloss} & ResNet-101 &39.1 &59.1 &42.3 &21.8 &42.7 &50.2 \\
				Faster R-CNN~\cite{frcnn} & ResNet-101 & 36.2 & 59.1 & 39.0 & 18.2 & 39.0 & 48.2 \\
				Deformable R-FCN~\cite{rfcn} & Inception-ResNet-v2 & 37.5 & 58.0 & 40.8 & 19.4 & 40.1 & 52.5 \\
				Mask R-CNN~\cite{maskrcnn} & ResNet-101 &  38.2 & 60.3 & 41.7 & 20.1 & 41.1 & 50.2 \\
		    SNIP~\cite{snip} & DPN-98 &45.7 &67.3 &51.1 &29.3 &48.8 &57. \\
				RefineDet512~\cite{refinedet}  & ResNet-101 &36.4 &57.5 &39.5 &16.6 &39.9 &51.4 \\
				Cascade R-CNN~\cite{cascadercnn} & ResNet-101 &42.8 &62.1 &46.3 &23.7 &45.5 &55.2 \\
				CornerNet~\cite{cornernet}  & Hourglass-104 &40.5 &56.5 &43.1 &19.4 &42.7 &53.9 \\
        CenterNet~\cite{centernet}   & Hourglass-104 &44.9 &62.4 &48.1 &25.6 &47.4 &57.4 \\
        RepPoints~\cite{reppoints} & ResNet-101-DCN &45.0 &66.1 &49.0 &26.6 &48.6 &57.5 \\
        GA-RPN~\cite{garpn} & ResNet-50 &39.8 &59.2 &43.5 &21.8 &42.6 &50.7 \\
        FoveaBox~\cite{kong2020foveabox} & ResNeXt-101 &42.1 &61.9 &45.2 &24.9 &46.8 &55.6 \\
        FCOS~\cite{fcos} & ResNeXt-64x4d-101 &43.2 &62.8 &46.6 &26.5 &46.2 &53.3 \\
        ATSS~\cite{atss} & ResNet-101 &43.6 &62.1 &47.4 &26.1 &47.0 &53.6 \\
        ATSS~\cite{atss} & ResNeXt-64x4d-101-DCN &47.7 &66.5 &51.9 &29.7 &50.8 &59.4 \\
		\midrule
				Libra R-CNN & ResNet-50             & 42.1 & 62.8 & 46.1 & 25.3 & 44.8 & 51.3  \\
				Libra R-CNN & ResNet-101            & 43.7 & 64.1 & 47.9 & 25.8 & 46.7 & 53.9  \\
				Libra R-CNN & ResNeXt-64x4d-101-DCN & 47.2 & 67.8 & 51.7 & 28.9 & 49.9 & 59.3  \\
		\bottomrule
	\end{tabular}
	\caption{Comparisons with state-of-the-art object detection methods on COCO \emph{test-dev}.}
	\label{tab:overall-results}
\end{table*}

\subsection{Experiments on MSCOCO}
We conduct experiments both on object detection and instance segmentation tasks on the challenging MS COCO~\cite{coco} benchmark.

\subsubsection{Dataset and Evaluation Metrics}
MS COCO consists of 115k images for training (\emph{train-2017}) and 5k images for validation (\emph{val-2017}).
There are also 20k images in \emph{test-dev} that have no disclosed labels.
We train models on \emph{train-2017}, and report ablation studies and final results on \emph{val-2017} and \emph{test-dev} respectively.
All reported results follow standard COCO-style Average Precision (AP) metrics that include
AP (averaged over IoU thresholds), AP$_{50}$ (AP for IoU threshold 50\%), AP$_{75}$ (AP for IoU threshold 75\%).
We also include AP$_{S}$, AP$_M$, AP$_L$, which correspond to the results on small, medium and large scales respectively.
The COCO-style Average Recall (AR) with AR$^{100}$, AR$^{300}$, AR$^{1000}$ correspond to the average recall when there are 100, 300 and 1000 proposals per image respectively.

\subsubsection{Implementation Details}
For fair comparisons, all experiments are implemented on PyTorch~\cite{pytorch} and mmdetection~\cite{mmdetection}.
The backbones used in our experiments are publicly available.
For ablation experiments, we train detectors with 8 GPUs (2 images per GPU) for 12 epochs with an initial learning rate of 0.02, and decrease it by 0.1 after 8 and 11 epochs respectively if not specifically noted.
All other hyper-parameters follow the settings in mmdetection~\cite{mmdetection} if not specifically noted.

\begin{table}[t]
	\centering
	\addtolength{\tabcolsep}{-5pt}
	\begin{tabular}{*{12}{c}}
		\toprule
		Method           & Backbone        & $\text{AP}$ & $\text{AP}_{50}$ & $\text{AP}_{75}$ \\
		\midrule
		RetinaNet$^*$~\cite{focalloss} & ResNet-50 & 35.8 & 55.3 & 38.6 \\
		Libra RetinaNet & ResNet-50 & 37.8 & 56.9 & 40.5 & \\
		\midrule
		\midrule
		Method           & Backbone        & $\text{AR}^{100}$ & $\text{AR}^{300}$ & $\text{AR}^{1000}$ \\
		\midrule
		RPN$^*$~\cite{frcnn}          & ResNet-50-FPN   & 42.5              & 51.2              & 57.1               \\
		RPN$^*$~\cite{frcnn}          & ResNet-101-FPN  & 45.4              & 53.2              & 58.7               \\
		RPN$^*$~\cite{frcnn}          & ResNeXt-101-FPN & 47.8              & 55.0              & 59.8               \\
		Libra RPN (ours) & ResNet-50-FPN   & \textbf{52.1}     & \textbf{58.3}   & \textbf{62.5}               \\
		\bottomrule
	\end{tabular}
	\caption{Comparisons on RetinaNet and RPN. The symbol ``*'' means our re-implements with $1 \times$ schedule in mmdetection.}
	\label{tab:rpn}
\end{table}

\begin{table*}[t]
	\centering
	\addtolength{\tabcolsep}{0pt}
	\begin{tabular}{*{12}{c}}
		\toprule
		IoU-balanced Sampling & Balanced Feature Pyramid & LGR-$L_1$ Loss & LGR-CE Loss & AP   & $\text{AP}_{50}$ & $\text{AP}_{75}$ & $\text{AP}_{S}$ & $\text{AP}_{M}$ & $\text{AP}_{L}$ \\
		\midrule
		                      &                          &    &  & 35.9 & 58.0             & 38.4             & 21.2            & 39.5            & 46.4            \\
		\checkmark            &                          &    &       & 36.8 & 58.0             & 40.0             & 21.1            & 40.3            & 48.2            \\
		\checkmark            & \checkmark               &    &       & 37.7 & 59.4             & 40.9             & 22.4            & 41.3            & 49.3            \\
		\checkmark            & \checkmark               & \checkmark    &   & 38.5 & 59.3             & 42.0             & 22.9            & 42.1            & 50.5            \\
		\checkmark            & \checkmark               & \checkmark    & \checkmark   & 39.0 & 60.0             & 42.6             & 23.1            & 42.5            & 50.7            \\
		\bottomrule
	\end{tabular}
	\caption{Effects of each component in our Libra R-CNN. Results are reported on COCO \emph{val-2017}.}
	\label{tab:overall-ablation}
\end{table*}

\subsubsection{Main Results on Object Detection}
We compare Libra R-CNN with the state-of-the-art object detection approaches on the COCO \emph{test-dev} subset.
For fair comparisons with recent detectors, we follow the practices in ~\cite{fcos,atss} to train the models.
That is, we randomly select a scale between 640 to 800 to resize the shorter side of the input images during training.
We double the training schedule to 24 epochs and decrease it by 0.1 after 16 and 22 epochs respectively.
For the bounding box head, we use the \emph{4conv-fc} head with group normalization~\cite{wu2018group} to keep consistent with~\cite{fcos,atss}.
We freeze all batch normalization layers in the backbone.
Inspired by ATSS~\cite{atss}, we also constrain the center point of proposals within the ground truth bounding boxes.

From Tabel~\ref{tab:overall-results}, we can observe that Libra R-CNN is on par with the recent states of the art.
Our method obtains 42.1 and 43.7 points AP with ResNet-50 and ResNet-101 backbone respectively.
The results with ResNet-101 is on par with the powerful ATSS~\cite{atss} on the overall AP and outperforms it by 2.0 points on mAP at IoU threshold of 0.5.
When training with the much powerful ResNeXt-64x4-101 backbone and deformable convolutions~\cite{dai2017deformable}, our method can obtain 47.2 points AP.
Although the result is slightly lower than ATSS by 0.5 points, our method still outperforms it by 1.3 points mAP at IoU threshold of 0.5.
These results show that our method is superior to ATSS on object classification but inferior to it on accurate localization.

Apart from the two-stage frameworks, we further extend our Libra R-CNN to single stage detectors and report the results of Libra RetinaNet.
Considering that there is no sampling procedure in RetinaNet~\cite{focalloss}, Libra RetinaNet only integrates balanced feature pyramid and balanced $L_1$ loss.
As shown in Table~\ref{tab:rpn}, without bells and whistles, Libra RetinaNet brings 2.0 points higher AP with ResNet-50 and achieves 37.8 AP.

Our method can also enhance the average recall of proposal generation.
As shown in Table~\ref{tab:rpn}, Libra RPN brings $9.2$ points higher AR$^{100}$, $6.9$ points higher AR$^{300}$ and $5.4$ points higher AR$^{1000}$ compared with RPN with ResNet-50 respectively.
Note that larger backbones only bring little gain to RPN.
Libra RPN can achieve 4.3 points higher AR$^{100}$ than ResNeXt-101-64x4d only with a ResNet-50 backbone.
The significant improvements from Libra RPN validate that the potential of RPN is much more exploited with the effective balanced training.

\begin{figure*}[t]
	\centering
	\includegraphics[width=\linewidth]{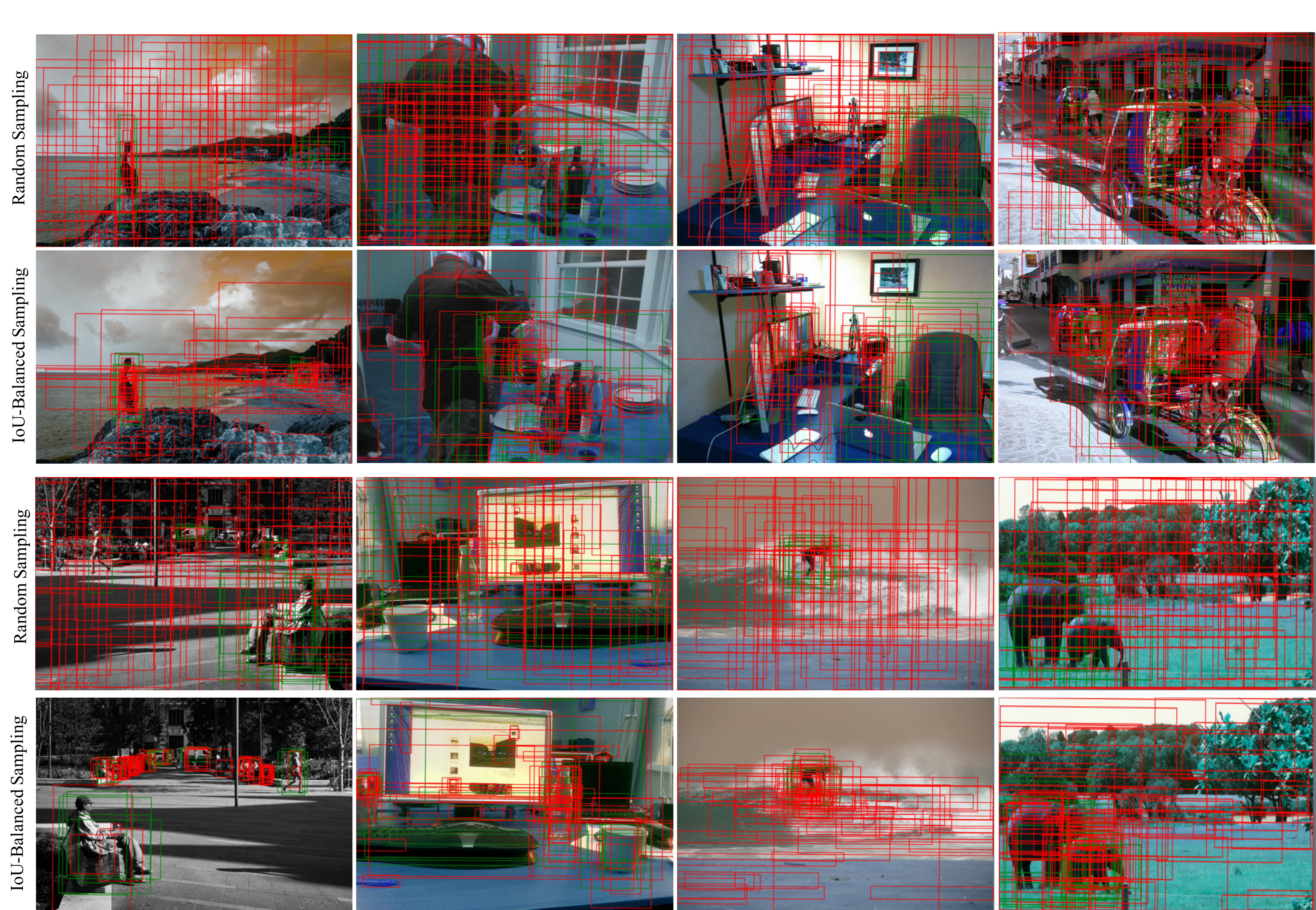}
	\caption{Visualization of training samples under random sampling and IoU-balanced sampling respectively.}
	\label{fig:sampling}\end{figure*}

\jm{
Because our method focuses on improving the training process of the detectors, we do not bring many extra costs during the inference time. 
When implementing Libra R-CNN with a ResNet-50 backbone, we can achieve an inference speed of 19.5 FPS on NVIDIA Tesla V100, while the speed of Faster R-CNN baseline is 20.4 FPS. 
This comparison shows that the improvements on the learning process can greatly enhance the model power without many extra costs.
}

\subsubsection{Ablation Experiments}

\paragraph{Overall Ablation Studies.}
To analyze the importance of each proposed component, we report the overall ablation studies in Table~\ref{tab:overall-ablation}.
We gradually add IoU-balanced sampling, balanced feature pyramid, LGR-$\text{L}_1$ loss, and LGR-CE Loss on ResNet-50 FPN Faster R-CNN baseline.
Experiments for ablation studies are implemented with the same pre-computed proposals for fair comparisons.

1) IoU-balanced Sampling:
IoU-balanced sampling brings 0.9 points higher box AP than the ResNet-50 FPN Faster R-CNN baseline, validating the effectiveness of this cheap hard mining method.
We also visualize the training samples under random sampling and IoU-balanced sampling in Figure~\ref{fig:sampling}.
It can be seen that the selected samples are gathered to the regions where we are more interested in instead of randomly appearing around the target.

2) Balanced Feature Pyramid:
Balanced feature pyramid improves the box AP from 36.8 to 37.7.
Results in small, medium and large scales are consistently improved, which validate that the balanced semantic features balanced low-level and high-level information in each level and yield consistent improvements.

3) Logarithmic Gradient Re-weighting:
LGR-$\text{L}_1$ loss improves the box AP from 37.7 to 38.5.
To be more specific, most of the improvements are from $AP_{75}$, which yields 1.1 points higher AP compared with corresponding baseline.
LGR-CE loss improves the box AP from 38.5 to 39.0.
The AP$_{50}$ and AP$_{75}$ are improved by 0.7 points and 0.6 points respectively. 
These results validate that the localization accuracy and classification capability of the model are much improved.

\begin{table}[t]
	\centering
	\addtolength{\tabcolsep}{-2pt}
	\begin{tabular}{*{12}{c}}
		\toprule
		Settings    & AP   & $\text{AP}_{50}$ & $\text{AP}_{75}$ & $\text{AP}_{S}$ & $\text{AP}_{M}$ & $\text{AP}_{L}$ \\
		\midrule
		Baseline    & 35.9 & 58.0             & 38.4             & 21.2            & 39.5            & 46.4            \\
		\midrule
		Pos Balance & 36.1 & 58.2             & 38.2             & 21.3            & 40.2            & 47.3            \\
		$K = 2$     & 36.7 & 57.8             & 39.9             & 20.5            & 39.9            & 48.9            \\
		$K = 3$     & 36.8 & 57.9             & 39.8             & 21.4            & 39.9            & 48.7            \\
		$K = 5$     & 36.7 & 57.7             & 39.9             & 19.9            & 40.1            & 48.7            \\
		\bottomrule
	\end{tabular}
	\caption{Ablation studies of IoU-balanced sampling on COCO \emph{val-2017}.}
	\label{tab:iou}
\end{table}

\begin{table}[t]
	\centering
	\addtolength{\tabcolsep}{-2pt}
	\begin{tabular}{*{12}{c}}
		\toprule
		Settings          & AP   & $\text{AP}_{50}$ & $\text{AP}_{75}$ & $\text{AP}_{S}$ & $\text{AP}_{M}$ & $\text{AP}_{L}$ \\
		\midrule
		Baseline          & 35.9 & 58.0             & 38.4             & 21.2            & 39.5            & 46.4            \\
		\midrule
		Integration       & 36.3 & 58.8             & 38.8             & 21.2            & 40.1            & 46.3            \\
		Refinement        & 36.8 & 59.5             & 39.5             & 22.3            & 40.6            & 46.5            \\
		\midrule
		PAFPN\cite{panet} & 36.3 & 58.4             & 39.0             & 21.7            & 39.9            & 46.3            \\
		Balanced PAFPN    & 37.2 & 60.0             & 39.8             & 22.7            & 40.8            & 47.4           \\
		\bottomrule
	\end{tabular}
	\caption{Ablation studies of balanced feature pyramid on COCO \emph{val-2017}.}
	\label{tab:fpn}
\end{table}

\begin{table}[t]
	\centering
	\addtolength{\tabcolsep}{-2pt}
	\begin{tabular}{*{12}{c}}
		\toprule
		Settings                        & AP   & $\text{AP}_{50}$ & $\text{AP}_{75}$ & $\text{AP}_{S}$ & $\text{AP}_{M}$ & $\text{AP}_{L}$ \\
		\midrule
		Baseline                        & 35.9 & 58.0             & 38.4             & 21.2            & 39.5            & 46.4            \\
		\midrule
		loss weight = 1.5               & 36.4 & 58.0             & 39.7             & 20.8            & 39.9            & 47.5            \\
		loss weight = 2.0               & 36.2 & 57.3             & 39.5             & 20.2            & 40.0            & 47.5            \\
		$L_1$ Loss (1.0)                   & 36.4 & 57.4 & 39.1 & 21.0 & 39.7 &  47.9\\
		$L_1$ Loss (1.5)                   & 36.6 & 57.2 & 39.8 & 20.2 & 40.0 & 48.2 \\
		$L_1$ Loss (2.0)                   & 36.4 & 56.5 & 39.6 & 20.1 & 39.8 & 48.2 \\
		\midrule
		$\alpha = 0.2$,  $\gamma = 1.0$ & 36.7 & 58.1             & 39.5             & 21.4            & 40.4            & 47.4            \\
		$\alpha = 0.3$,  $\gamma = 1.0$ & 36.5 & 58.2             & 39.2             & 21.6            & 40.2            & 47.2            \\
		$\alpha = 0.5$,  $\gamma = 1.0$ & 36.5 & 58.2             & 39.2             & 21.5            & 39.9            & 47.2            \\
		\midrule
		$\alpha = 0.5$,  $\gamma = 1.5$ & 37.2 & 58.0   & 40.0             & 21.3            & 40.9            & 47.9            \\
		$\alpha = 0.5$,  $\gamma = 2.0$ & 37.0 & 58.0             & 40.0             & 21.2            & 40.8            & 47.6            \\
		\bottomrule
	\end{tabular}
	\caption{Ablation studies of balanced $L_1$ loss on COCO \emph{val-2017}. The numbers in the parentheses indicate the loss weight.}
	\label{tab:loss}
\end{table}

\begin{table}[t]
	\centering
	\addtolength{\tabcolsep}{-2pt}
	\begin{tabular}{*{12}{c}}
		\toprule
				Settings                        & AP   & $\text{AP}_{50}$ & $\text{AP}_{75}$ & $\text{AP}_{S}$ & $\text{AP}_{M}$ & $\text{AP}_{L}$ \\
		\midrule
		Libra R-CNN~\cite{librarcnn} & 38.5 & 59.3             & 42.0             & 22.9            & 42.1            & 50.5            \\
		\midrule

		$\alpha = 0.5$,  $\gamma = 1.0$ & 38.7 & 59.6             & 42.4             & 22.9            & 42.2            & 50.4              \\
		$\alpha = 0.8$,  $\gamma = 1.0$ & 39.0 & 60.0             & 42.6             & 23.1            & 42.5            & 50.7             \\
		$\alpha = 1.0$,  $\gamma = 1.0$ & 38.8 & 59.8             & 42.3             & 23.0            & 42.2            & 50.5             \\
		\bottomrule
	\end{tabular}
	\caption{Ablation studies of LGR-CE Loss on COCO \emph{val-2017}.}
	\label{tab:loss2}
\end{table}

\begin{figure*}[t]
	\centering
	\includegraphics[width=\linewidth]{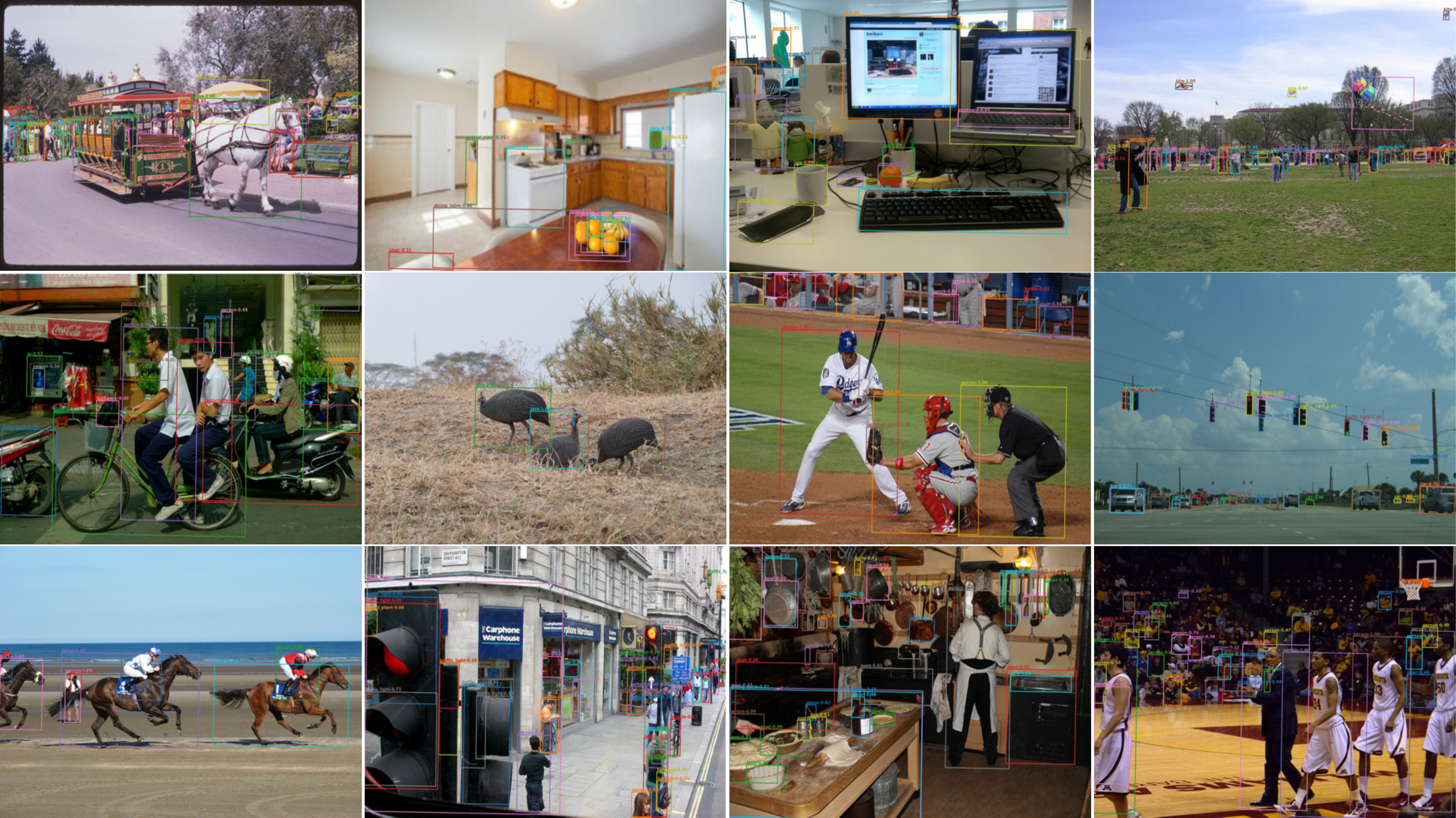}
	\caption{Visualizations of object detection results with Libra R-CNN on MS COCO.}
	\label{fig:visual_det}
\end{figure*}

\paragraph{Ablation Studies on IoU-balanced Sampling.}
Experimental results with different implementations of IoU-balanced sampling are shown in Table~\ref{tab:iou}.
We first verify the effectiveness of the complementary part, \ie sampling equal number of positive samples for each ground truth, which is stated in Section~\ref{subsec:sample} and denoted by \emph{Pos Balance} in Table~\ref{tab:iou}.
Since there are too few positive samples to explore the potential of this method, this sampling method provides only small improvements (0.2 points higher AP) compared to ResNet-50 FPN Faster R-CNN baseline.

Then we evaluate the effectiveness of IoU-balanced sampling for negative samples with different hyper-parameters $K$, which denotes the number of intervals.
Experiments in Table~\ref{tab:iou} show that the results are very close to each other when the parameter $K$ is set as 2, 3 or 5.
Therefore, the number of sampling interval is not much crucial in our IoU-balanced sampling, as long as the hard negatives are more likely selected.

\paragraph{Ablation Studies on Balanced Feature Pyramid.}
Ablation studies of balanced feature pyramid are shown in Table~\ref{tab:fpn}.
We also report the experiments with PAFPN~\cite{panet}.
We first implement balanced feature pyramid only with integration.
Results show that the naive feature integration brings 0.4 points higher box AP than the corresponding baseline.
Note there is no refinement and no parameter added in this procedure.
With this simple method, each resolution obtains equal information from others.
Although this result is comparable with the one of PAFPN~\cite{panet}, we reach the feature level balance without extra convolutions, validating the effectiveness of this simple method.

Along with the embedded Gaussian non-local attention~\cite{nonlocal}, balanced feature pyramid can be further enhanced and improve the final results.
Our balanced feature pyramid is able to achieve 36.8 AP on COCO dataset, which is 0.9 points higher AP than ResNet-50 FPN Faster R-CNN baseline.
More importantly, the balanced semantic features have no conflict with PAFPN.
Based on the PAFPN, we include our feature balancing scheme and denote this implementation by Balanced PAFPN in Table~\ref{tab:fpn}.
Results show that the Balanced PAFPN is able to achieve 37.2 box AP on COCO dataset, with 0.9 points higher AP compared with the PAFPN.

\paragraph{Ablation Studies on Balanced $L_1$ Loss.}
Ablation studies of balanced $L_1$ loss are shown in Table~\ref{tab:loss}.
We observe that the localization loss is mostly half of the recognition loss.
Therefore, we first verify the performance when raising loss weight directly.
Results show that tuning loss weight only improves the result by 0.5 points.
And the result with a loss weight of 2.0 starts to drop down.
These results show that the outliers bring a negative influence on the training process, and leave the potential of model architecture from being fully exploited.
We also conduct experiments with $L_1$ loss for comparisons.
Experiments show that the results are inferior to ours.
Although the overall results are improved, the AP$_{50}$ and AP$_S$ drop obviously.

In order to compare with tuning loss weight directly, we first validate the effectiveness of balanced $L_1$ loss when $\gamma = 1$.
Balanced $L_1$ loss is able to bring 0.8 points higher AP than baseline.
With our best setting, balanced $L_1$ loss finally achieves $37.2$ AP, which is 1.3 points higher than the ResNet-50 FPN Faster R-CNN baseline.
These experimental results validate that our balanced $L_1$ achieves a more balanced training and makes the model better converged.

\begin{figure*}
	\centering
	\includegraphics[width=\linewidth]{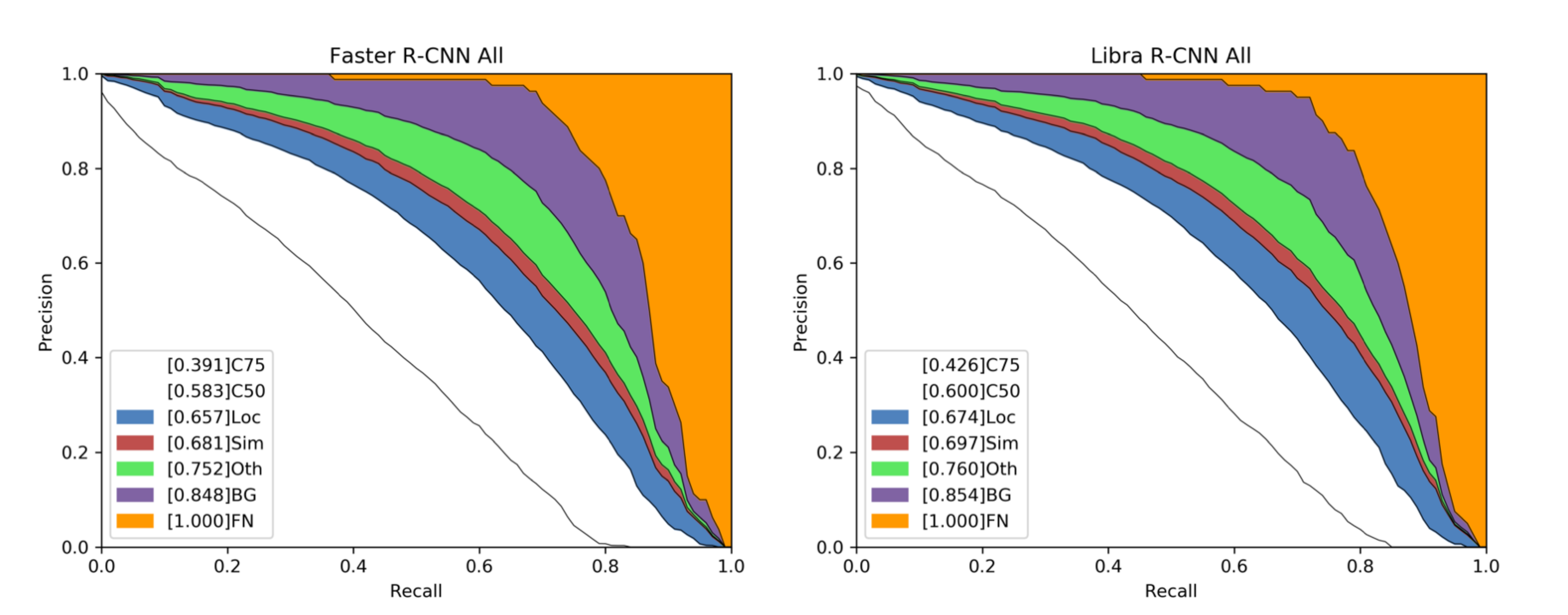}
	\caption{Object detection error analysis of Faster R-CNN and Libra R-CNN following~\cite{deterroranalysis}.}
	\label{fig:analysis}
\end{figure*}
\begin{table*}[htb]
	\centering
	\resizebox{\linewidth}{!}{
	\begin{tabular}{c|c|c|cccccc|cccccc}
		\toprule
		Method                       & Backbone  & Schedule   &  $AP^\text{M}$ & $\text{AP}_{50}^\text{M}$ & $\text{AP}_{75}^\text{M}$ & $\text{AP}_{S}^\text{M}$ & $\text{AP}_{M}^\text{M}$ & $\text{AP}_{L}^\text{M}$ & $AP^\text{B}$ & $\text{AP}_{50}^\text{B}$ & $\text{AP}_{75}^\text{B}$ & $\text{AP}_{S}^\text{B}$ & $\text{AP}_{M}^\text{B}$ & $\text{AP}_{L}^\text{B}$ \\
		\midrule
		Mask R-CNN~\cite{maskrcnn}& ResNet-50-FPN & $1\times$ & 34.1 & 55.4 & 36.2 & 18.1 & 37.5 & 46.4 & 37.1 & 58.7 & 40.1 & 22.0 & 40.7 & 47.8            \\
		\midrule
		Libra Mask R-CNN (ours) & ResNet-50-FPN & $1\times$ & 35.3 & 57.2 & 37.5 & 19.0 & 38.3 & 48.1 & 39.5 & 60.5 & 43.2 & 23.3 & 42.9 & 50.9 \\
		\bottomrule
	\end{tabular}
	}
	\caption{Experimental results of Libra Mask R-CNN on MS COCO \emph{val-2017}.
		The ``$1\times$'' training schedules follow the settings explained in mmdetection~\cite{mmdetection}.}
	\label{tab:overall-results2}
\end{table*}

\paragraph{Ablation Studies on LGR-CE Loss.}
Ablation studies of LGR-CE loss are shown in Table~\ref{tab:loss2}.
For clear comparison with the original Libra R-CNN~\cite{librarcnn}, we directly set Libra R-CNN as the baseline.
We implement LGR-CE loss with different parameters and all found improvements.
When $\alpha = 0.8$, we found that the LGR-CE loss brings 0.5 points improvements, which proves the effectiveness of the method.
With LGR-CE loss, Libra R-CNN finally achieves $39.0$ AP, which is 3.0 points higher than the ResNet-50 FPN Faster R-CNN baseline.
These experimental results validate that our LGR-CE loss boosts the classification capability of the network.

With the overall balanced design, Libra R-CNN obtains promising detection results on MS COCO as shown in Figure~\ref{fig:visual_det}.

\subsubsection{Error Analysis}

Inspired by the method proposed by Derek Hoiem \etal~\cite{deterroranalysis} that diagnoses errors in object detection,
we further analyze the errors both in Faster R-CNN and Libra R-CNN.
The detailed analysis between Libra R-CNN and Faster R-CNN can help us better explore the sources of the improvements.
The methods analyze 7 terms: C75, C50, Loc, Sim, Oth, BG and FN.
C75, C50, and Loc represent the Precision-Recall curves with a IoU threshold of 0.75, 0.50, and 0.10 respectively.
Sim, Oth, BG, FN represent the Precision-Recall curves that errors with super-category false positives (fps),
all class confusions, background (and class confusion) fps, all remaining errors are removed respectively.

As shown in Figure~\ref{fig:analysis}, Libra R-CNN improves 3.5 points C75 and 1.7 points C50 compared to Faster R-CNN.
The improvement of C75 is much higher than the one of C50, which proves that the balanced designs greatly enhance the accurate localization capability of the network.
As the LGR is designed to further boost the well-localized and well-classified samples, this result further proves the effectiveness of the method.
From the analysis above we can observe that the improvements are gradually decreased.
The gains are mainly from C75 and Sim, which represents the accurate localization capability and cross-class classification capability respectively.
The observations prove that the region classifier and regressor are optimized better.

Although recent advancements have greatly pushed forward the states of the art, the overall performance is still unsatisfactory.
From Figure~\ref{fig:analysis}, we can observe that the classification capability is still the main block for the performance of instance recognition. Removing the localization errors can boost the performance by 7.4 points, while the remaining 32.6 points all rely on accurate classifications. 
These problems will become even worse when transferring to the long-tailed cases, which will be discussed below. 
It shows that current methods still struggle from the long-tailed datasets due to the unsatisfactory classification results.
It reveals that the researchers shall explore how to boost the power of the classifiers accordingly.

\begin{figure*}[htb]
	\centering
	\includegraphics[width=\linewidth]{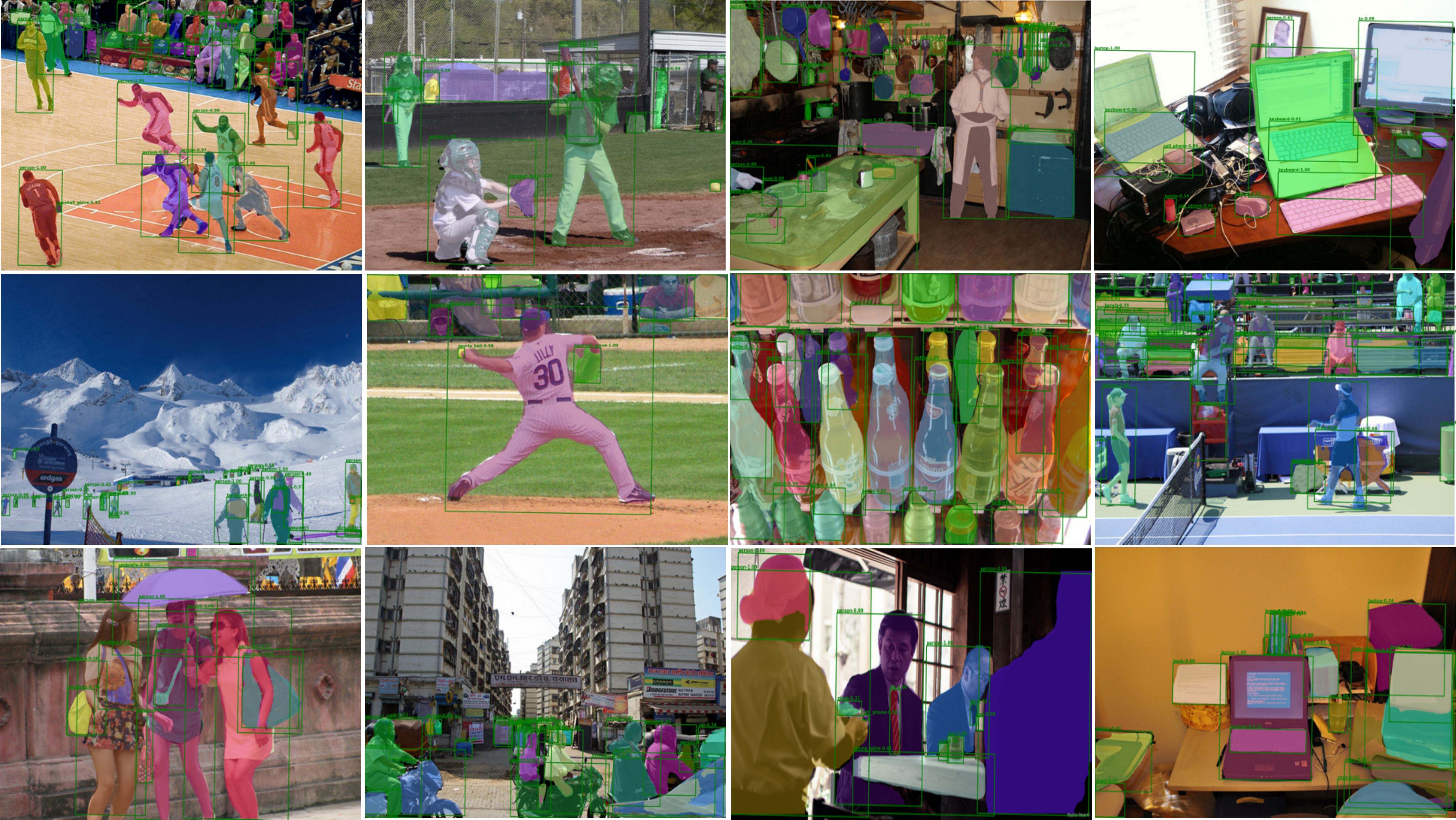}
	\caption{Visualizations of instance segmentation with Libra Mask R-CNN on MS COCO.}
	\label{fig:seg}
\end{figure*}

\begin{table*}[htb]
	\centering
	\begin{tabular}{c|c|c|ccccccc}
		\toprule
		Method                       & Set & BCR &  $AP$ & $\text{AP}_{50}$ & $\text{AP}_{75}$ & $\text{AP}_{r}$ & $\text{AP}_{c}$ & $\text{AP}_{f}$ & $\text{AP}^\text{b}$ \\
		\midrule
		Mask R-CNN~\cite{lvis}  & \emph{val}      & - & 22.0 & 35.0 & 23.4 & 10.7 & 21.2 & 27.8 & 22.3 \\
		Mask R-CNN~\cite{lvis}  & \emph{test-dev} & - & 22.0 & 34.9 & 23.1 & 10.2 & 21.6 & 27.9 & -    \\
		\midrule
		Ours  & \emph{val}      &  -          &  23.3 & 36.2 & 24.7 & 14.5 & 22.4 & 27.8 & 23.9  \\
		Ours  & \emph{val}      &  \checkmark &  25.0 & 38.4 & 26.5 & 16.8 & 25.0 & 28.1 & 25.8  \\
		Ours  & \emph{test-dev} &  \checkmark &  24.5 & 38.7 & 25.9 & 15.7 & 24.6 & 28.4 & -      \\
		\bottomrule
	\end{tabular}
	\caption{Experimental results on LVIS dataset. All results are implemented with repeat factor sampling in ~\cite{lvis}.}
	\label{tab:overall-lvis}
\end{table*}

\subsubsection{Results on Instance Segmentation}
Considering object detection and instance segmentation are inherently associated after the introduction of Mask R-CNN~\cite{maskrcnn},
we further conduct experiments on Mask R-CNN to see how the balanced design on object detectors improving the downstream pixel-wise segmentation performance.
We follow the implementation details in Section 4.2 and insert a mask branch to Libra R-CNN, like how Mask R-CNN differs from Faster R-CNN.

Experimental results are shown in Table~\ref{tab:overall-results2}.
Libra Mask R-CNN obtains 35.3 mask AP and 39.5 bbox AP on COCO \emph{val-2017} benchmark.
The improvements are 1.2 points and 2.4 points on bbox and mask respectively.
For bbox AP, the improvements on small, medium and large scale objects are 1.3 points, 2.2 points and 3.1 points respectively.
The conclusions are the same as stated in Section 4.4.
The results prove that the method works well for the detection branch in Mask R-CNN.

For mask AP, the improvements are lower than bbox AP, but still 1.2 points.
0.9 points, 0.8 points and 1.7 points mask AP are obtained for small, medium and large objects respectively.
The improvements of mask AP are mainly from the improvements of detection results.
The results prove that better locations of the bounding boxes, better the mask predictions.

The promising visualization results of Libra Mask R-CNN on MS COCO are shown in Figure~\ref{fig:seg}.

\subsection{Experiments on LVIS}

\begin{table}[t]
	\centering
	\begin{tabular}{*{12}{c}}
		\toprule
		Method    & AP   & $\text{AP}_{50}$ & $\text{AP}_{75}$ & $\text{AP}_{r}$ & $\text{AP}_{c}$ & $\text{AP}_{f}$ \\
		\midrule
		Baseline & 22.0 & 35.0 & 23.4 & 10.7 & 21.2 & 27.8 \\
		EQL~\cite{tan2020equalization} & 23.5 & 36.9 & 25.1 & 14.5 & 23.7 & 27.6 \\
		\midrule
		\text{t} = 0.05   &  23.5 & 36.8 & 25.0 & 14.7 & 23.5 & 27.5 \\
		\text{t} = 0.025  &  24.0 & 37.4 & 25.4 & 16.5 & 23.8 & 27.5 \\
		\text{t} = 0.01   &  23.6 & 37.0 & 25.2 & 16.0 & 23.3 & 27.4 \\
		\midrule
		\jm{\text{S} = 5}  &  23.9 & 37.5 & 25.4 & 16.3 & 23.6 & 27.4 \\
		\jm{\text{S} = 10}  &  24.0 & 37.4 & 25.4 & 16.5 & 23.8 & 27.5 \\
		\jm{\text{S} = 15}   &  24.0 & 37.4 & 25.3 & 16.6 & 23.9 & 27.5 \\
		\bottomrule
	\end{tabular}
	\caption{Ablation studies of bucketed class re-weighting on LVIS \emph{val} subset.}
	\label{tab:ab:bcr}
\end{table}

We further conduct experiments on LVIS~\cite{lvis} dataset to verify the effectiveness of our efforts on class-level balancing.

LVIS~\cite{lvis} is recently proposed for large vocabulary instance segmentation which contains 1203 classes for training and testing.
It has 100k images for training~(\emph{train}), 20k images for validation~(\emph{val}), and 20k images for testing~(\emph{test-dev}).
We train the models on \emph{train} and report the results on \emph{val} and \emph{test-dev} respectively.
We report results with Average Precision (AP) across different IoU thresholds, AP$_{50}$ (AP for IoU threshold at 0.50), AP$_{75}$ (AP for IoU threshold at 0.75).
We also include AP$_r$, AP$_c$, and AP$_f$ which represent AP for rare, common and frequent classes respectively.
The APs for bounding boxes are also included on the validation set. 

We train the models following the $1 \times$ schedule in mmdetection~\cite{mmdetection}.
We randomly select a scale between 640 to 800 to resize the shorter side of the input images during training.
The output threshold for objects is set as 0.001 to keep consistent with the original paper~\cite{lvis}.
All models are implemented with the repeat factor sampling with a hyper-parameter of 0.001 following~\cite{lvis}.
The other hyper-parameters all keep consistent with the aforementioned ones on MSCOCO if not mentioned.

The results of our method are shown in Table~\ref{tab:overall-lvis}.
Our method obtains 25.0 and 24.5 points AP on LVIS \emph{val} and \emph{test-dev} subsets, brining 3.0 points and 2.5 points improvements to the Mask R-CNN baseline respectively.
On the \emph{val} subset, the AP of rare classes is greatly improved from 10.7 to 16.8 that prove the effectiveness for class-level balancing.
We also conduct ablation experiments on BCR and present the results in Table~\ref{tab:ab:bcr}. We can observe that the best result outperforms the Mask R-CNN baseline by 2.0 points on overall AP and 5.8 points on rare classes.
The result also outperforms the recent work EQL~\cite{tan2020equalization} by 0.5 points on overall AP and 2.0 points on rare classes. 
This comparison greatly proves the effectiveness of our method on class-level balancing.

It is worth noting that the original Libra R-CNN design without the bucketed class reweighing can bring 1.3 points and 3.8 points AP improvements on overall performance and rare classes respectively. 
The BCR further improves the overall AP by 1.7 points and the AP of rare classes by 2.3 points. 
Therefore, the large improvement of the original Libra R-CNN design on rare classes proves that the balancing design on the other aspects can also benefit the class-level balancing.

\begin{table}[t]
	\centering
	\begin{tabular}{*{12}{c}}
		\toprule
		Method   & Backbone & Input size & mAP            \\
		\midrule
		Faster R-CNN~\cite{frcnn} & VGG-16     & $\sim 1000 \times 600$ &  73.2 \\
		Faster R-CNN~\cite{frcnn} & ResNet-101 & $\sim 1000 \times 600$ & 76.4 \\
    Faster R-CNN$^*$ & ResNet-50           & $\sim 1000 \times 600$ & 80.9 \\
    R-FCN~\cite{rfcn} & ResNet-101         & $\sim 1000 \times 600$ & 80.5 \\
    YOLOv2~\cite{yolo9000} & DarkNet-19    & $\sim 544 \times 544$ & 78.6 \\
    RefineDet~\cite{refinedet} & VGG-16    & $\sim 512 \times 512$ & 82.5 \\
    Kong~\etal~\cite{kong2018deep} & ResNet-101 & $\sim 512 \times 512$ & 82.4 \\
    \midrule
    Ours            & ResNet-50  & $\sim 1000 \times 600$ & 84.3  \\
    Ours             & ResNet-101 & $\sim 1000 \times 600$ & 85.4  \\
		\bottomrule
	\end{tabular}
	\caption{Results on Pascal VOC dataset. The symbol ``*'' means our re-implements.}
	\label{tab:voc}
\end{table}

\subsection{Experiments on Pascal VOC}

We conduct experiments on Pascal VOC~\cite{voc} dataset to verify the generalization ability of proposed methods.
Pascal VOC dataset includes 20 classes for training and testing.
We train all models on the union set of the VOC 2007 and 2012 \emph{trainval} sets, and evaluate them on the VOC 2007 \emph{test} set, which is the common practice in ~\cite{frcnn,ssd,refinedet}.
We directly use the same hyper-parameters on MS COCO to train the models.
During the training process, we resize the input images to keep their shorter side being 600 and their longer side less or equal to 1000.
We train detectors with 8 GPUs (2 images per GPU) for 12 epochs with an initial learning rate of 0.01, and decrease it by 0.1 after 9 epochs.
We report the results of mean Average Precision (mAP) with a IoU threshold of 0.5.
The results are shown in Table~\ref{tab:voc}.

For a fair comparison, we re-implement Faster R-CNN ResNet-50 as the baseline and implement our methods based on Faster R-CNN.
We can observe that our method outperforms the baseline by 3.4 points.
Our method is superior to recent works such as RefineDet~\cite{refinedet} and Kong~\etal~\cite{kong2018deep} by ~3.0 points.
The results prove the generalization ability of our method that it could be extended to other datasets without tuning hyper-parameters.

\section{Conclusion}
In this paper, we systematically revisit the training process of detectors,
and find the potential of model architectures is not fully exploited due to the imbalance issues existing in the training process.
Based on the observation, we propose Libra R-CNN to balance the imbalance through an overall balanced design.
With the help of the simple but effective components, Libra R-CNN brings significant improvements on MS COCO, LVIS, and Pascol VOC datasets.
Extensive experiments show that Libra R-CNN well generalizes to various backbones for both two-stage detectors and single-stage detectors.

\paragraph{Acknowledgement}
This work is partially supported by
the Science and Technology Plan of Zhejiang Province of China (No. 2017C01033),
the Civilian Fundamental Research (No. D040301),
the Collaborative Research grant from SenseTime Group (CUHK Agreement No. TS1610626 \& No. TS1712093),
and the General Research Fund (GRF) of Hong Kong~(No. 14236516 \& No. 14203518).


%
%

\bibliographystyle{spmpsci}      
\bibliography{egbib}   


\end{document}